\documentclass{article}

\usepackage{arxiv}

\usepackage[utf8]{inputenc} 
\usepackage[T1]{fontenc}    
\usepackage{hyperref}       
\usepackage{url}            
\usepackage{booktabs}       
\usepackage{amsfonts}       
\usepackage{nicefrac}       
\usepackage{microtype}      
\usepackage{lipsum}
\usepackage{graphicx}
\graphicspath{ {./figs/} }

\newcommand{\smallheader}[1]{\noindent\textit{#1}\par}

\title{Time Series Analysis of Urban Liveability}

\author{
 Alex Levering\\
  Wageningen University\\
  Wageningen, the Netherlands \\
  \texttt{alex.levering@wur.nl} \\
   \And
Diego Marcos \\
  INRIA\\
  Montpellier, France \\
  \And
 Devis Tuia \\
  EPFL\\
  Sion, Switzerland \\
}

\begin{document}
\maketitle

\begin{abstract}
In this paper we explore deep learning models to monitor longitudinal liveability changes in Dutch cities at the neighbourhood level. Our liveability reference data is defined by a country-wise yearly survey based on a set of indicators combined into a liveability score, the Leefbaarometer. We pair this reference data with yearly-available high-resolution aerial images, which creates yearly timesteps at which liveability can be monitored. We deploy a  convolutional neural network trained on an aerial image from 2016 and the Leefbaarometer score to predict liveability at new timesteps 2012 and 2020. The results in a city used for training (Amsterdam) and one never seen during training (Eindhoven) show some trends which are difficult to interpret, especially in light of the differences in image acquisitions at the different time steps. This demonstrates the complexity of liveability monitoring across time periods and the necessity for more sophisticated methods compensating for changes unrelated to liveability dynamics.
\end{abstract}

\section{Introduction}
Cities are dynamic and complex entities with many different and competing functions. One such function is the provision of living space to its residents, which is commonly referred to as the liveability of the city. It is commonly defined as \textit{"the degree to which its provisions and requirements fit with the needs and capacities of its members"} \cite{veenhoven_happiness_1993}. It therefore promotes the human-centric design of cities such that they are made into places where people want to live. A failure to provide liveable environments can result in a variety of detrimental health effects, such as higher mortality rates \cite{haan_poverty_1987} and higher morbidity rates \cite{barber_neighborhood_2016}. As such, there is a vested interest from policymakers to provide liveable environments for residents.

Measuring liveability is generally a costly and time-consuming process as it is done through the surveying of residents. Recent research has attempted to predict liveability at a larger scale by means of remote sensing imagery \cite{suel_multimodal_2021,levering_predicting_2023}. Such models are now delivering promising results which potentially allow for monitoring without requiring the acquisition of expensive reference data. The next step in operationalizing liveability monitoring is to explore the potential for such models to predict over multiple timesteps. If such models can reliably forecast liveability, then data-constrained countries may consider the adoption of remote sensing for monitoring the liveability of the territory.

In this paper we discuss the potential for deep learning on aerial images to monitor the liveability of cities across multiple time steps. We discuss the creation of a liveability monitoring dataset over The Netherlands and test the applicability of pre-trained models for liveability estimation to demonstrate the need for more sophisticated time-series modelling approaches accounting for domain shifts related to different acquisition conditions, camera, etc. We study multi-temporal liveability estimation in two Dutch cities and use the Leefbaarometer, a Dutch project providing longitudinal studies for liveability in the Netherlands, as reference data to evaluate our strategy.

\section{Liveability Monitoring data}
Assessing the liveability of cities has been an actively studied topic over the years, with more recent literature focusing on predictive modeling. Recent work has made use of advances in deep learning models to predict liveability factors directly from overhead imagery, proving that liveability factors can be accurately predicted with remote sensing \cite{suel_multimodal_2021, scepanovic_jane_2021}. However, liveability monitoring requires multiple timesteps of both the reference data for neighbourhoods and the aerial images, and for this reason there have thus far been limited research efforts which attempt to monitor liveability dynamics (i.e. longitudinally over time) from overhead images. In the following, we detail our two sources of data, which we used for monitoring at three time steps: 2012, 2016 (data used to train the model) and 2020.

\subsection{Reference data: the Leefbaarometer.}
\begin{figure*}[!t]
    \centering
    \includegraphics[scale=0.35]{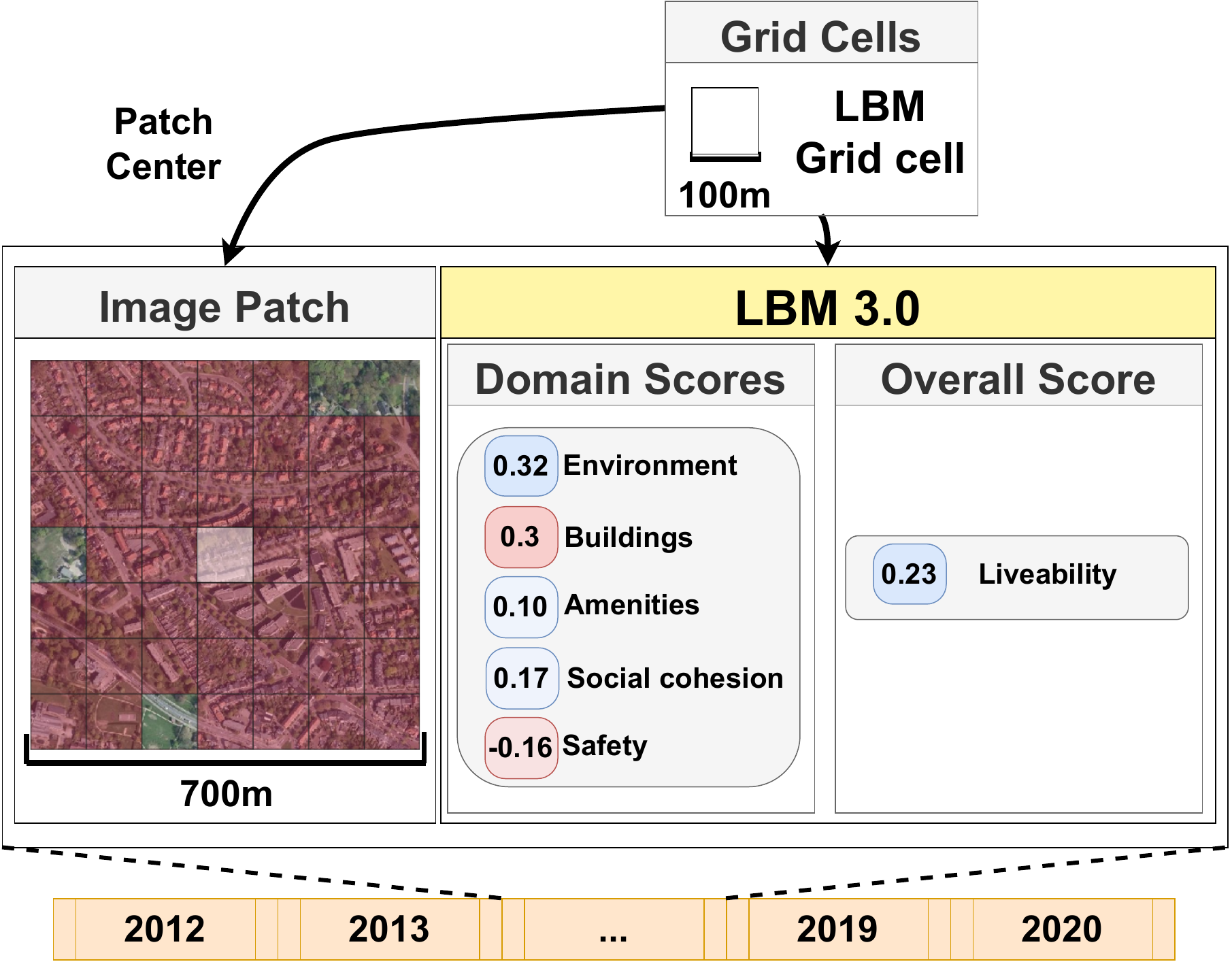}
    \caption{Data construction process for the liveability time series monitoring task. For each year there is a grid with liveability reference data available, which can be used to match the yearly aerial images to the neighbourhood liveability data. Each grid cell also has liveability domain score information available, which is the contribution of each domain to the liveability of a neighbourhood. As it is zero-centered, the overall score of 0.23 indicates that this patch is above averagely liveable.}
    \label{fig:ranks_2020}
\end{figure*}

The Netherlands is a notable exception to the lack of reference data. The Dutch government  has monitored the liveability of residents for many years through a project known as the \emph{Leefbaarometer} (LBM). In the LBM project, researchers attempt to model the opinion of residents on their liveability as a function of various publicly available datasources which can be grouped as \emph{liveability domains}. These domains encompass factors which affect liveability, such as socio-economic variables, housing quality, and access to amenities. The LBM project therefore measures both how liveable a neighbourhood is, as well as how various domains contribute to that liveability. The 3.0 version of the LBM project released in 2022 reaches an $R^{2}$ of 0.75 in the reproduction of the residents responses with the five selected domains (center column of Fig.~\ref{fig:ranks_2020}). Furthermore, the LBM is verified to be accurate by policy makers. In this version, the LBM model is calculated yearly from 2012 until 2020, and is due to be updated for the recent and coming years. In this work we use the liveability scores provided by LBM for each 100 m $\times$ 100 m inhabited cell across the Netherlands.

\subsection{Aerial data.}
We use very high resolution aerial imagery to predict the liveability scores. Since 2016, the government of Netherlands releases a yearly cloud-free winter aerial image with a uniform sensor and 25cm pixel resolution across the entire country. Further yearly images are available before this timestep as well, with a lower resolution and fewer timesteps. The continued availability of both data sources means that the amount of data for this monitoring task will continue to increase over the years. Fig.~\ref{fig:ranks_2020} illustrates a set of image patches, together with the Liveability score and their  intermediate domain scores. To encourage the use of the dataset, we provide scripts and links to the data to prepare it in the  following Github repository: \hyperlink{https://github.com/ahlevering/LBMOverheadMonitor}{https://github.com/ahlevering/LBMOverheadMonitor}.

\section{Monitoring using Pre-trained Models}
We developed a machine learning model to predict liveability, as well as the domain scores, in~\cite{levering_predicting_2023} and we use that model as a starting point to study the complexity of monitoring  multiple timesteps. For this purpose we use a pre-trained model to predict the liveability at new timesteps. Our model is trained on an aerial image from 2016 (details in~\cite{levering_predicting_2023}) and we apply it without modifications to aerial images from 2012 and 2020. For both timesteps, we compare the overall liveability score with the ground truth provided by the LBM 3.0 project. While our model is trained on the 2.0 version of the LBM project, there is compatibility with the 3.0 version as they both measure liveability as the deviation to the national average. However, to maintain compatibility, we resize the image to 500x500 pixels, rather than 600x600. For this paper we do not consider the intermediate sub-scores  as the definition of the sub-scores changed between versions 2.0 and 3.0. We assess neighbourhood shifts relative to one another by analyzing their relative ranking within the city. If one neighbourhood improves then, by comparison, another neighbourhood may seem less attractive. This also makes the prediction case less sensitive to mean shifts when predicting on new images. We use the reference data from the LBM 3.0 to determine if the model picks up on measured shifts in the liveability of neighbourhoods.

We consider two cities for our experiment chosen for their contrasting qualities, namely Amsterdam and Eindhoven. Amsterdam is the biggest conurbation of The Netherlands. It has a long history and it is unique in its design compared to other cities in the country. Eindhoven is an industrial city which urbanized only since the industrial revolution. Further, it important to note that Amsterdam was included in the training cities in used to build the model of~\cite{levering_predicting_2023}, while Eindhoven was included in its test set and was never seen during model training.

\section{Results and discussion}
\subsection{Liveability rankings}
\begin{figure*}[!t]
    \centering
    \includegraphics[scale=0.72]{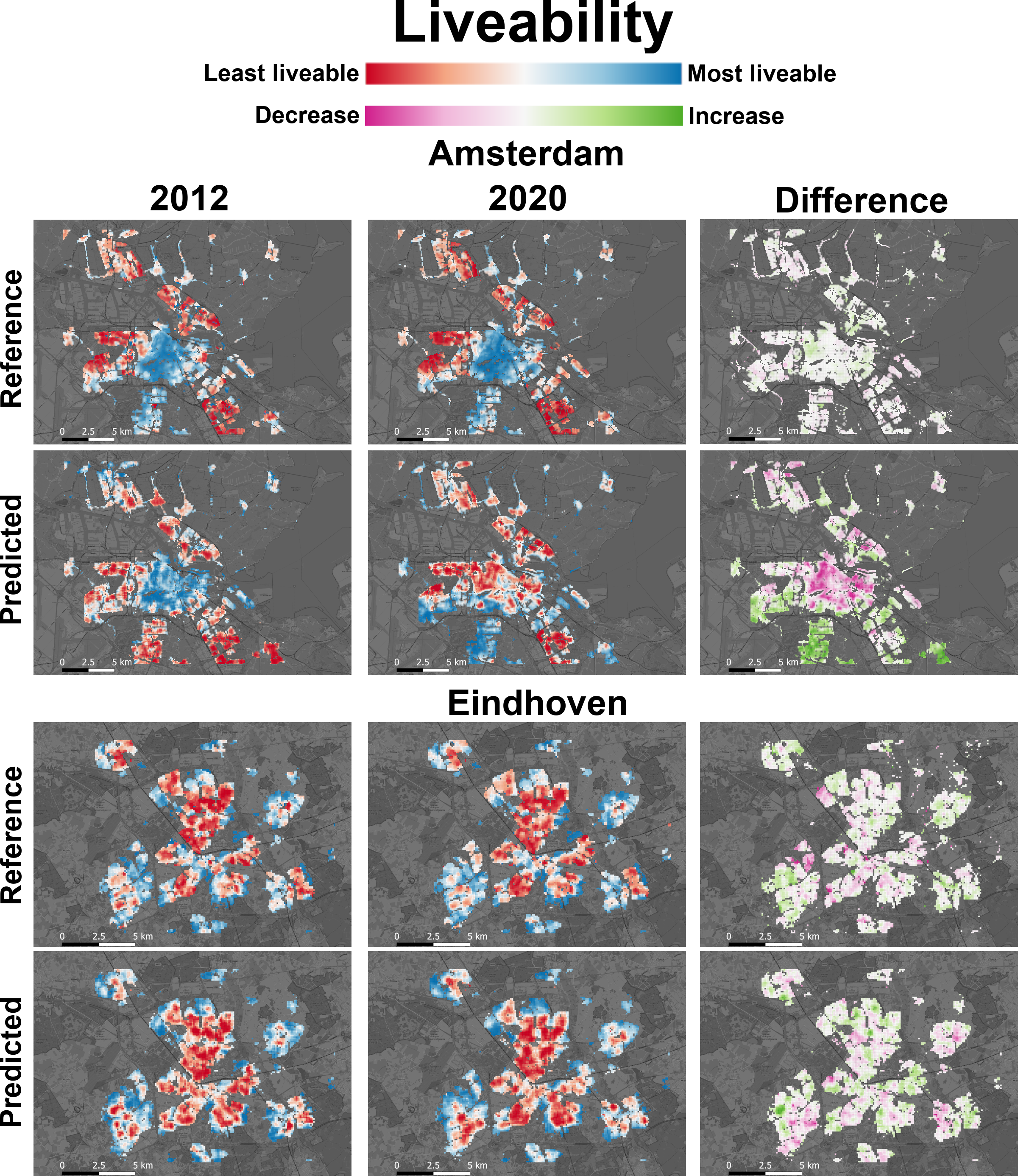}
    \caption{Liveability rankings as predicted by a pre-trained liveability prediction model compared to the LBM 3.0 reference values. Neighbourhoods are ranked from lowest (red) to highest (blue) ranking. We also show the changes between 2012 and 2020 from purple (negative change) to green (positive change) for both cities.}
    \label{fig:changes}
\end{figure*}

In this section we discuss how our model predicted the liveability rankings in Amsterdam and Eindhoven for two new timesteps unseen during training. Firstly we present Kendall's $\tau$ for both 2012 and 2020 in table \ref{tab:results}. $\tau$ is a  coefficient calculating correlations between in two lists using ranking of indices rather than the actual values of such indices. The sharp decrease in performance on the new timesteps in Amsterdam suggests that model has overfitted to this region during training. However, the minor decrease in accuracy over Eindhoven suggests that liveability prediction models can remain stable when predicting new timesteps, even in geographical regions not seen during training. In Fig.~\ref{fig:changes} we show the predicted scores, the LBM 3.0 reference values, and the changes between 2012 and 2020. We discuss the patterns for each city separately.

\begin{table}[h]
\centering
\caption{Kendall's $\tau$ of the predictions compared to the LBM 3.0 reference scores. We also include the comparison to 2016, which is the year on which the model was trained.}
\begin{tabular}{l|ccc}
\textbf{}          & \textbf{2012 $\tau$} & \textbf{2016 $\tau$} & \textbf{2020 $\tau$} \\ \hline
\textbf{Amsterdam} & 0.316 & 0.656 & 0.337                       \\
\textbf{Eindhoven} & 0.487 & 0.537 & 0.478              
\label{tab:results}
\end{tabular}
\end{table}

\vspace{.3cm}\smallheader{Amsterdam}
The predictions at different timesteps for Amsterdam show some correct change patterns, but a closer inspection yields various issues. At timestep 2012, the model predicts the overall trends correctly, but with some notable exceptions: it especially mispredicts the liveability of the peripheral areas. Most notably, the model ranks \emph{Amstelveen} (south-western peripheral city) among the worst areas to live. In turn, it considers areas in the north of Amsterdam to be more liveable than the reference data. The model's predictions for the 2020 reference data are even more unexpected. Firstly, the model is better able to predict the liveability of the periphery where it follows the overall trend better. However, in turn it severely underpredicts the liveability of the city center of Amsterdam. As a whole, the 2020 prediction shows swings in liveability which are unrealistic given the time difference to the year in which the original model was trained (2016). This result reveals that the predictive qualities of liveability prediction models can break down when introduced with new time series data. Our results corroborate previous research which suggest that liveability factors of culturally distinct cities are more difficult to generalize \cite{scepanovic_jane_2021}. Moreover, improving predictions for the center of Amsterdam through normalization experiments resulted in the degradation of predictions in other cities, which indicates that there is a domain shift problem which needs to be studied in more detail. 

\clearpage
\smallheader{Eindhoven}
For the city of Eindhoven the results are less erratic over the years. Eindhoven is part of the test set in our original study~\cite{levering_predicting_2023}, so it is surprising that the model has more stable ranking predictions here compared to Amsterdam. For both timesteps, the model predicts fewer strong swings in liveability, and generally predicts changes in the same areas as the reference data. However, the changes it predicts in some places trend in the wrong direction. For instance, near \emph{Veldhoven}, which is the bottom left block of grid cells, it correctly predicts that the north of the village has a positive development trend. However, the model does not predict an overall improvement in the entire area, as it considers the south of the village to be less liveable in 2020. It is an encouraging result that the model is able to provide stable ranking predictions which fluctuate in the same areas as the reference data, but more research is needed to provide models which can more accurately correctly predict the direction of these changing trends.

\subsection{Liveability monitoring}
With the release of the third version of the LBM and the availability of aerial images both at yearly resolution (2012 onwards), it is now possible to approach this problem as a time series prediction problem. The results above show some pointers to inspire future research efforts in liveability modeling with deep learning. First, our results suggest that such a task is not trivial, and will likely require a more sophisticated approach. Approaches rooted in deep domain adaptation~\cite{wang_deep_2018} seem to be necessary, for example to reduce the gap in the image space between the different years acquisitions for example proposing image normalisation strategies that would align the years: fine-tuning on the specific year data, projective methods~\cite{gross_nonlinear_2019, tasar_colormapgan_2020} or adversarial approaches~\cite{kellenberger_deep_2021} are potential options to be explored, to be sure that the discrepancies between monitoring years are due to changes in liveability, rather than differences in acquisition conditions of the images.

\section{Conclusions}
In this paper we have studied the task of liveability monitoring from aerial imagery over the entirety of The Netherlands and focused on the examples of the cities of Amsterdam and Eindhoven. We created and made available a dataset from the yearly liveability reference data and aerial images, which opens up the potential for long-term monitoring of liveability. We have also tested the applicability for pre-trained liveability prediction models to predict over new timesteps, which has shown promising results for the general trends, but also highlighted the necessity for more advanced methods for this task able to account for domain differences and provide accurate liveability maps out of the training domain.

\bibliographystyle{unsrt}

\end{document}